\documentclass{article}

\usepackage{arxiv}

\usepackage[utf8]{inputenc} 
\usepackage[T1]{fontenc}    
\usepackage{hyperref}       
\usepackage{url}            
\usepackage{booktabs}       
\usepackage{amsfonts}       
\usepackage{nicefrac}       
\usepackage{microtype}      
\usepackage{float} 
\usepackage{graphicx} 
\usepackage{amsmath} 
\usepackage{multicol} 
\usepackage{framed,multirow} 
\title{Landmark Guidance Independent Spatio-channel Attention and Complementary Context Information based Facial Expression Recognition}
\author{
  Darshan Gera \\
  \texttt{darshangera@sssihl.edu.in} \\
   \And
S Balasubramanian \\
  \texttt{sbalasubramanian@sssihl.edu.in} \\
}

\begin{document}
\maketitle

\begin{abstract}
A recent trend to recognize facial expressions in the real-world scenario is to deploy attention based convolutional neural networks (CNNs) locally to signify the importance of facial regions and, combine it with global facial features and/or other complementary context information for performance gain. However, in the presence of occlusions and pose variations,  different channels respond differently, and further that the response intensity of a channel differ across spatial locations. Also, modern facial expression recognition(FER) architectures rely on external sources like landmark detectors for defining attention. Failure of landmark detector will have a cascading effect on FER. Additionally, there is no emphasis laid on the relevance of features that are input to compute complementary context information. Leveraging on the aforementioned observations,  an end-to-end architecture for FER is proposed in this work that obtains both local and global attention per channel per spatial location through a novel spatio-channel attention net (SCAN), without seeking any information from the landmark detectors. SCAN is complemented by a complementary context information (CCI) branch. Further, using efficient channel attention (ECA), the relevance of features input to CCI is also attended to. The representation learnt by the proposed architecture is robust to occlusions and pose variations. Robustness and superior performance of the proposed model is demonstrated on both in-lab and in-the-wild datasets (AffectNet, FERPlus, RAF-DB, FED-RO, SFEW, CK+, Oulu-CASIA and JAFFE) along with a couple  of constructed face mask datasets resembling masked faces in COVID-19 scenario. Codes are publicly available at \url{https://github.com/1980x/SCAN-CCI-FER}.
\end{abstract}

\section{Introduction} \label{Introduction}
It is rightly said that face is the index of mind. The emotional states (calm, happy, anger, sad, fear, disgust, surprise, contempt) of a person can be understood from the facial expressions of the person. In communication and social-interaction, facial expression and body language, together with words, make the interaction more meaningful. In analysing depression and mental related illness, longitudinal study of facial expressions play a vital role. In student-teacher interaction, facial expression can indicate attentiveness, teaching quality etc.

With the advancement in technology, there has been a continuous upsurge in the interest of researchers to enable machines to recognize facial expressions. This  field of research is called as facial expression recognition (FER). Machines with FER capability has plethora of applications viz human-computer interaction \cite{1}, intelligent tutoring systems \cite{27}, automatic driver alert systems \cite{2} , mental health analysis \cite{3}, support systems for mentally retarded children \cite{4}, computer animation \cite{5} etc.

A way to enable machines to perform FER is to make them learn the facial expressions through examples. Traditional works like \cite{6}, \cite{7}, \cite{8}, \cite{9} focused on training machines for FER through examples collected in a controlled (in-lab) environment. Examples of such in-lab datasets are CK+ \cite{5}, Oulu-CASIA \cite{11} and JAFFE \cite{44}. However, machines trained in controlled environment do not generalize well to real-world scenarios wherein the facial images contain occlusions, vary in illumination, differ in pose etc. To tackle these challenges, large datasets in the order of millions of facial images like AffectNet \cite{12} collected from real-world scenarios (in-the-wild) have been made available. Despite this, developing algorithms that are robust to occlusion, pose and illumination variation are not trivial. For example, the object of occlusion could be internal (for eg., hands occluding the face) or external (for eg., a face mask or glasses), size of occluder could vary and that any location in the image could be occluded. Similarly, extreme pose variations like pose in profile facial images could misinform the algorithm due to loss of symmetry. Illumination variation can block information proportional to the degree of illumination. Misinformation or loss of information is heavy on algorithms as they have to search for appropriate regions containing relevant information, which could be minimal in size, to discriminate well between different facial expressions. This is where attention mechanism pitches in.

Attention mechanism attempts to weigh relevant regions (for eg., non-occluded regions) heavier for further processing than less relevant regions. This enhances the discriminating power of the algorithm for FER. Local attention is particularly useful in this regard, especially in the presence of occlusions. In local attention, local patches or regions of the input are focused upon. Their relevance  is subsequently quantified through the attention mechanism. Consideration of local patches make identification of relevant regions relatively easier since, for example,  the object of occlusion could cover certain local  patches, thereby making them less relevant eventually as the machine learns. Local attention is generally combined with global context obtained from the undivided attention weighted input for performance gain.

With the introduction of graphics processing units (GPUs) and since the renaissance of neural networks through AlexNet \cite{13} under the deep learning (DL) umbrella, deep convolutional neural networks (CNNs) have super-scaled the performance in recognition problems \cite{13}, \cite{14}, \cite{15}, \cite{16}. An important factor that drives the success of CNNs and in fact any deep network is the large volume of data. With huge face datasets like MS-Celebrity face dataset \cite{17}, Casia webface \cite{18}, VGGFace \cite{19}, VGGFace2 \cite{20} etc. for face recognition(FR), the current trend for FER is to transfer knowledge (weights) from pre-trained FR  models to FER model and then fine-tune the weights on FER datasets like AffectNet. Further, to account for occlusions, pose and illumination variations, attention mechanism is incorporated in to the CNNs and the complete model is fine-tuned \cite{21}, \cite{22}, \cite{23}. Specifically, local attention weighted features are combined with complementary context information like global facial features for FER. This trend has significantly improved the recognition rate in FER, for example from around 50\% \cite{28} to around 60\% \cite{23} on AffectNet dataset though there is still a wide scope for improvement.

A particular issue has been with regard to how the attention mechanism is incorporated into the CNN. A very recent work \cite{29} has illustrated that in presence of wild scenarios like presence of occlusions, that are common in real-world scenes, different channels of a CNN feature map respond differently to the input stimuli, and further that the response intensity of a channel differ across spatial locations. However, attention based CNNs \cite{21}, \cite{22}, \cite{23} for FER have either applied constant attention weight across both channel and spatial dimensions, or applied per channel attention weight that is constant across spatial dimensions, or per position attention weight that is constant across channel dimensions. Further, some of these attention based CNNs \cite{21}, \cite{23} define local attention based on external sources like landmark detectors \cite{24}, \cite{25}. Erratic behaviour of such sources will naturally pull down the performance, and such anomalous behaviour is possible under challenging scenarios of presence of occlusions and pose and illumination variations. Also, these attention based CNNs do not emphasize on the relevance of features input to the part of the model that extracts complementary context information to local attention module.

Leveraging on these observations,  an end-to-end architecture for FER is proposed in this work that obtains both local and global attention per channel per spatial location, without seeking any information from the landmark detectors. Further,  using efficient channel attention (ECA) \cite{26}, the relevance of features input to compute  complementary context information is also attended to. The proposed architecture is robust to occlusion and pose variation. The branch of the proposed architecture that computes  attention is called as spatio-channel attention net (SCAN). The other branch, that again operates on facial regions, is called as complementary context information (CCI) branch. The input to both the branches is the high-level semantic representation of the face in the given facial image, obtained from a backbone network (ResNet-50) that was pre-trained on a large face dataset (VGGFace 2). The details of the proposed architecture will be presented in Section 3. In summary our contributions are as follows:

\begin{itemize}
    \item A local-global attention branch called as SCAN that computes attention weight per channel per spatial location  manner across all considered local patches and the whole input to make FER model robust to occlusion and pose variation. We also illustrate that the local attention branch can be shared across all local patches without any significant performance dip. This makes the local attention branch lightweight. 
    \item An enhanced complementary context information branch called CCI, enhanced by ECA, that provides richer complementary information to SCAN.
    \item An end-to-end architecture for FER devoid of dependence on external sources like landmark detectors to define attention.
    \item Validation of the proposed architecture on variety of datasets, both in-lab and in-the-wild cases with superior performance over the recent methods.  Additionally, comparison against a baseline and state of the art method on a couple of constructed masked-facial datasets resembling masked faces in COVID-19 scenario.
\end{itemize}
\section{Related Work}
Earlier works relied on hand-crafted features like deformable features, motion features and statistical features along with traditional machine learning (ML) paradigms like SVM for FER \cite{30}, \cite{31}, \cite{32}. These works validated their models on in-lab datasets \cite{10_a}, \cite{11}, \cite{44}. However, models trained for controlled environment do not generalize well outside that environment. Current trend is being driven by DL  paradigm with availability of huge volume of data collected in an uncontrolled ‘in-the-wild’ setting. It is to be noted that in-the-wild datasets, though available for FER, are not upscaled yet except for AffectNet \cite{12}. However, as mentioned earlier, datasets for FR are very huge. So, a general strategy for FER is to adopt transfer learning \cite{34} from FR.

CNNs have stamped their supremacy with regard to performance on recognition problems. In fact, almost two decades ago, CNNs have been found to be robust to some degree of affine transformations with regard to FER \cite{33}. In \cite{34}, a two-stage architecture is proposed for FER. In the first stage, parameters of FR model guides the learning of convolutional layers of the FER model. In the subsequent stage,  expression representation is learnt by addition of fully connected (fc) layers, and further refinement. \cite{35} addresses FER through an ensemble network with a novel pre-processing transformation, that transforms image intensities into a 3d-space, facilitating robustness to illumination variation. While \cite{35} used an ensemble network, \cite{36} uses an ensemble of supervisions through supervised scoring ensemble (SSE) learning mechanism wherein, apart from the output layer, intermediate and shallow layers are also supervised. Spatial information from facial expression images and differential information from changes between neutral and non-neutral expressions have been fused together using multiple fusion strategies in \cite{37} for FER. Multi-scale features obtained using hybrid inception-residual units with concatenated rectified linear activations enabled the architecture in \cite{38} to capture variations in expression. A 2-channel CNN with one channel trained in an unsupervised fashion, and subsequent information merge from both the channels aid FER in \cite{39}. Apart from stand-alone CNNs, generative models where CNNs define the generator and discriminator have also been used for FER. In \cite{40}, a facial image is viewed as a sum of neutral and expressive component, and subsequently the residual expressive component is learnt using a de-expression residual learning module under a generative model setting. Another generative model that suppreses identity while preserving expression, to account for within subject variance, is proposed in \cite{41}. While all these methods outperform traditional hand-crafted features ML based methods, most of them still validate their analysis only on in-lab datasets. They are not robust to challenges like presence of occlusions and pose variations that are common in ‘in-the-wild’ datasets. Incorporation of attention mechanism plays a vital role to face these challenges in FER.

Attention mechanism, an attempt to imitate how humans focus on salient regions in an image, has been successful in many computer vision tasks including abnormal behaviour recognition \cite{42} and visual explanation \cite{43}. With regard to FER robust to occlusions and pose variations, the recent works \cite{21}, \cite{22}, \cite{23} have pushed up the performance. In \cite{21}, unobstructedness or importance scores of local patches of feature maps corresponding to certain landmark points are computed using self attention and the respective local feature maps are weighted by these scores. The expectation is that, over training, patches corresponding to occluded areas in the image will receive low importance and hence become less relevant. Parallely, global context is captured through self attention on the entire feature map. Concatenation of local-global context is passed to a classifier for expression recognition. Region attention network (RAN) \cite{22} is conceptually similar to \cite{21} but selects patches directly from the image input. RAN combined with a region biased loss quantifies the importance of patches. Subsequently, a relation-attention module that relates local and global context provides the expression representation for classification. It is to be noted that selection of patches directly from the image input will increase time of inference. In \cite{23}, attention weights are generated as samples from a Gaussian distribution centered at spatial locations of the feature map, corresponding to certain confident landmark points, where the confidence score is provided by an external  landmark detector. Selection of local patches follow \cite{21}. Concurrently, complementary information is gathered from non-overlapping partitions of the feature map. Together, the patch based information and the complementary information guide the classifier to report state-of-the-art results.\newline

This work is also an attention based architecture for robust FER. It differs from \cite{21}, \cite{22}, \cite{23} in the following aspects:
\begin{itemize}
    \item \cite{21} assigns a single attention weight to the entire feature map per patch. \cite{22} assigns attention weight to every feature per input crop but it works with a flattened vector thereby losing spatial information. \cite{23} assigns spatially varying attention weights that are constant across channel dimensions, per patch. In this work, per channel per spatial location attention weight per patch is defined through the proposed SCAN.
    \item \cite{22},\cite{23} uses landmark detector to define attention. The proposed model does not need external sources like landmark detectors to define attention.
    \item \cite{23} does not emphasize on the relevance of input feature maps to collect complementary information. The proposed work uses ECA to attend to the relevance of input feature maps to CCI branch.
    \item \cite{22} fixes patches at the level of image input, thereby potentially increasing the inference time, especially when the number of patches are scaled up. The proposed work does not crop the input into multiple patches.
\end{itemize}
\section{Proposed Model}
\subsection{Motivation}
\begin{figure}[htb!]
    \centering
    \includegraphics[width=.5\textwidth]{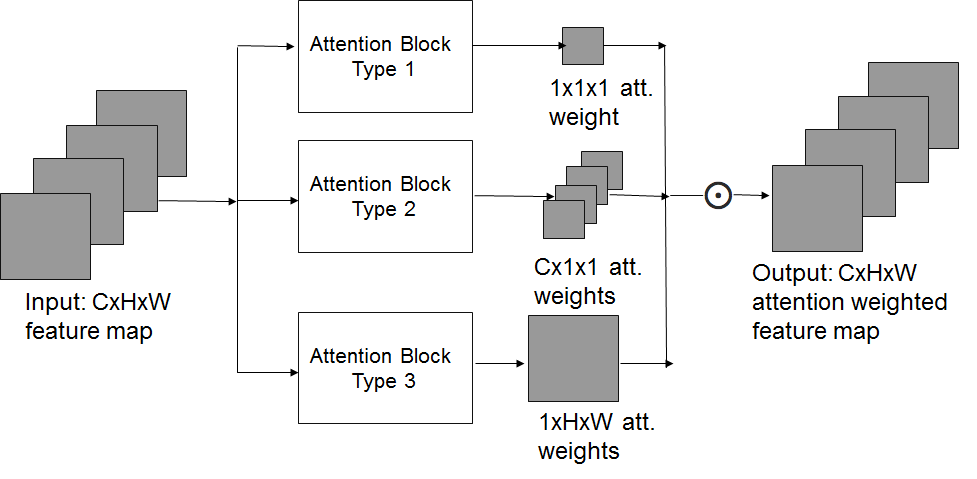}   
    \caption{Generic description of attention mechanism in \cite{21}, \cite{22}, \cite{23}}
    \label{fig:fig_attention_types}
\end{figure}

\begin{figure}[ht!]
    \centering
    \includegraphics[width=.5\textwidth]{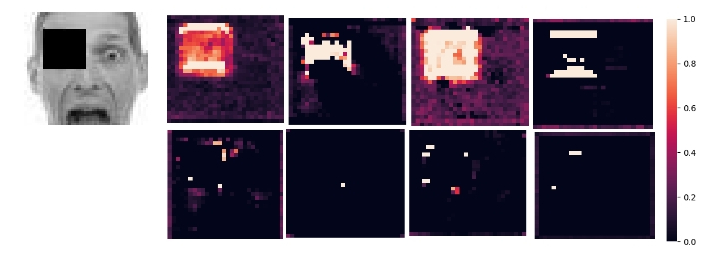}
    \caption{Channel response to occlusion. Top leftmost is a sample of occluded image. Other 8 images show heatmap of median relative change in response of 8 channels across a sample of 100 clean and occluded pairs.}
    \label{fig:figure_heatmaps}
\end{figure}

The proposed model is motivated by the attention based FER detailed in \cite{21}, \cite{22}, \cite{23} and the key observation in \cite{29}. A generic depiction of types of attention blocks in these attention based models is shown in Fig. \ref{fig:fig_attention_types}.
\newline
The three types of attention blocks, in order, are adopted in \cite{21}, \cite{22}, \cite{23} respectively. It can be noticed from Fig. \ref{fig:fig_attention_types} that either the attention weight is constant across spatial and channel dimensions (Type 1, \cite{21}) or it is constant across spatial dimensions (Type 2, \cite{22}) or it is constant across channel dimensions (Type 3, \cite{23}). However, it has been illustrated in \cite{29}, that in presence of occlusions, different channels respond differently and further that, each channel exhibits variation in intensity across spatial locations, for the given input stimuli. 
\newline
For example, Fig.\ref{fig:figure_heatmaps} shows 8 chosen channels of the median relative difference in response between a clean image and its occluded counterpart from the output of Conv\_3x block in ResNet-50. Median is computed from a sample of 100 pairs of clean and occluded images with occlusion fixed at the same location across the entire sample. Clearly, some channels display minimal response to occlusion while others show differing responses around the region of occlusion. 
\subsection{Spatio-Channel Attention Net(SCAN)}
\begin{figure}[htb!]
    \centering
    \includegraphics[width=.5\textwidth]{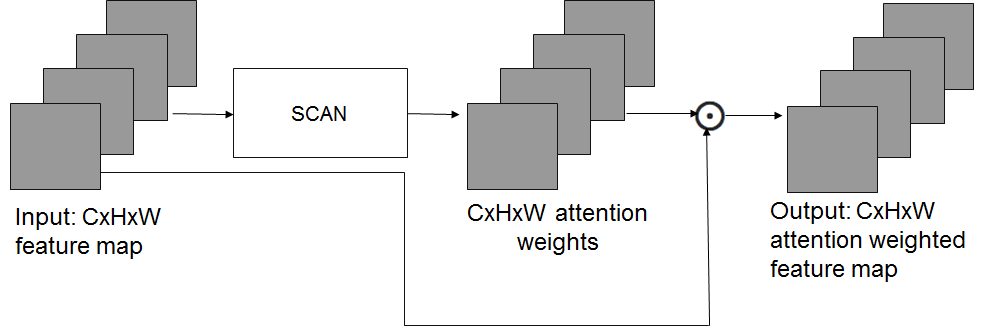}
    \caption{Input output pipeline of SCAN}
    \label{fig:figure_attention_net}
\end{figure}

Based on the aforementioned observation, this work proposes an attention block that provides attention per channel per spatial location for the given input feature map. This attention block is called as Spatio-Channel Attention Net. The input-output pipeline of SCAN is shown in Fig. \ref{fig:figure_attention_net}. Mathematically, let $P$  be the  $C\times H\times W$ dimensional input feature map to SCAN. Let $f$ be the non-linear function that models SCAN. In this work, $f$ is defined to be a ‘same’ convolution operation with number of out channels same as that of input channels, followed by parametric ReLU (PReLU) activation, batch normalization (BN) and a sigmoid activation ($\sigma$). Let $OP$ be the attention weights of the input feature map computed by SCAN. Then
\begin{equation}\label{eq:1}
    OP = SCAN(P) = f(P) = \sigma(BN(PReLU(Conv(P))))
\end{equation}      
where Conv is the ‘same’ convolution operation that takes a $C\times H\times W$ input feature map and outputs $C\times H\times W$ attention weights.

Following the computation of attention weights, the input feature map is weighted. Let $WP$ denote the weighted input feature map. Then,
\begin{equation}\label{eq:2}
     WP =  P \odot OP
\end{equation}
where $\odot$ is the element-wise multiplication operator. For sake of convenience, it is assumed that this element-wise multiplication is merged with SCAN. Hence the output of SCAN is $WP$ for the input $P$. It is to be noted that the input $P$ could be a local patch from a feature map, or the entire feature map providing global context. For this reason, SCAN is called as local-global attention branch. 

When multiple patches from a feature map are input to SCAN, the learnable parameters in SCAN can either be shared across all the input patches, or kept independent across patches. In this work, both shared as well as independent  set of SCAN parameters reported similar performance. This will be illustrated in Section \ref{Resultsanddiscussion}. For further presentation, it is assumed that parameters are independent across patches. 

\subsection{Complementary Context Information (CCI) Branch}
While SCAN attends to relevant features locally and globally, extra complementary information to this would surely enhance the discriminating ability of the model. Since transfer learning from FR is adopted for FER, the wealth of information available in FR could be easily leveraged upon to extract complementary context information. Particularly, information through feature maps from middle layers of the FR model could be used for this purpose. Middle layers are likely to contain features surrounding parts of the face like eyes, nose, mouth etc. that are useful for FER. Final layers generally contain identity specific features which are not suitable for FER. 

Features from the middle layer may be redundant, and all the features may not be important for FER. To eliminate redundancy and emphasize the important features, ECA \cite{26} is applied to the feature map from middle layer. ECA is a local cross-channel interaction mechanism without dimensionality reduction that computes channel-wise attention, constant across spatial dimensions. ECA has shown that dimensionality reduction  decreases the performance while cross-channel interaction improves performance. It is to be noted that this philosophy is followed in SCAN as well.
A natural question here would be that why ECA here and why not SCAN. It is to be noted that the goal here is to obtain complementary context information to the one provided by SCAN. Hence SCAN is not used here. Also, prior to using the information from the middle layer of FR model to perform FER, the important features needs to be emphasized and the  redundancy needs to be eliminated. Hence channel-wise attention suffices here. In this regard, since ECA has demonstrated its superiority over other attention methods, it is adopted here.

The CCI branch proposed in this paper is similar to the one in \cite{23} but with prior feature emphasis by ECA. The influence of ECA is illustrated in Section \ref{subsection_CCI_ECA}.  The CCI branch is described as follows:
Let F be the feature map from a chosen middle layer of the base model trained for FR. Let OF be the output of CCI branch. Then,
\begin{equation}  \label{eq:3}
      \begin{split}
            OF &= CCI(F) = GAP(PARTITION(ECA(F))) \\
               &= \{ OF_{1}, OF_{2}, ..., OF_{k}  \} 
      \end{split}
\end{equation}
where PARTITION partitions the ECA weighted feature map into k non-overlapping blocks and GAP is global average pooling. $OF_{i}$ is the output of GAP on $i^{th}$ block and it is a vector of features. Following \cite{23}, k is chosen to be 4 in this work.

\subsection{Proposed Architecture}
\begin{figure*}[ht!]
    \centering
    \includegraphics[width=1.0\textwidth]{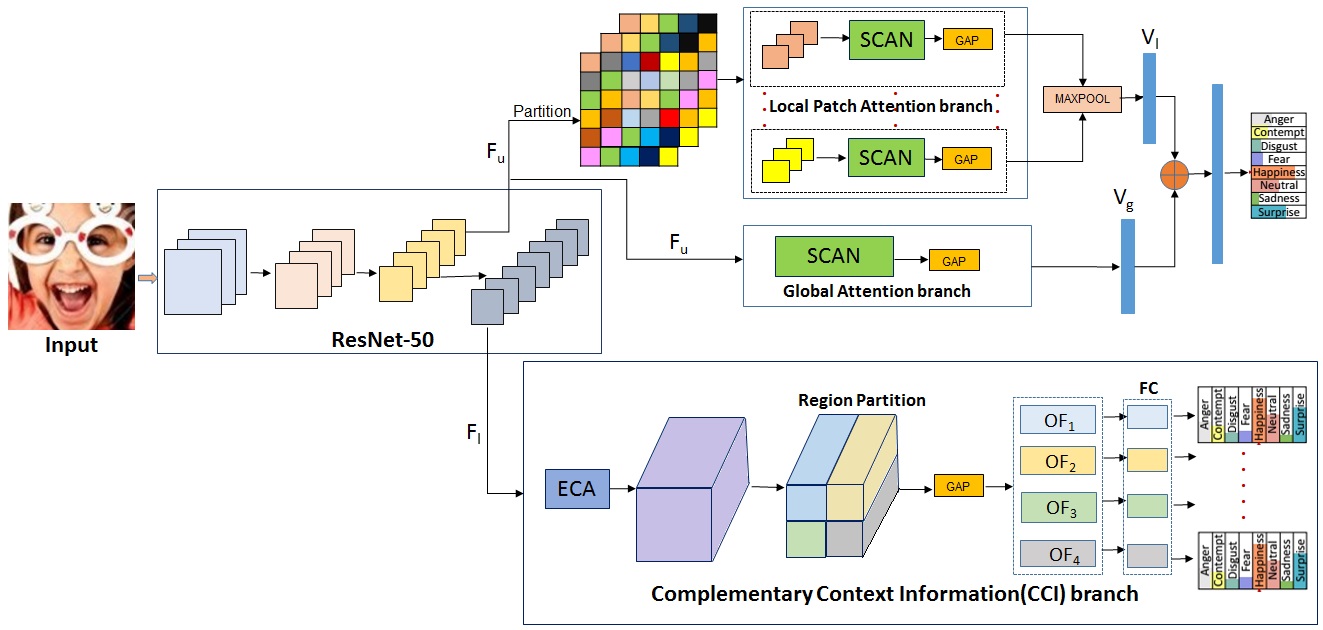}
    \caption{Complete Architecture of proposed Attention based Facial Expression Recognition Framework. For SCAN, refer Fig. \ref{fig:figure_attention_net} (Best viewed in color)}
    \label{fig:figure_attention_fraework}
\end{figure*}

The complete proposed architecture is shown in Fig. \ref{fig:figure_attention_fraework}.
The input I is a  $224 \times 224$  RGB image. It is processed by a pre-trained ResNet-50 model. For the SCAN branch, the output of $Conv\_3x$ block from ResNet-50 goes as input. This is a $28 \times 28 \times 512$ feature map, denoted by $F_{u}$. Next, $F_{u}$ is partitioned into m non-overlapping blocks, each block acting as a local patch. In this work, m is chosen to be 25 based on the ablation study reported in Section \ref{subsec_ablation_studies}. This results in 25 local patches. The output of SCAN from each of the 25 local patches is global average-pooled across spatial dimensions and subsequently max-pooled across channel dimensions to provide the summary of the local context in the form of a 512 dimensional vector, denoted by $V_{l}$. Also, the whole $F_{u}$ is fed to SCAN and the output is global average-pooled to capture global context in the form of another 512 dimensional vector, denoted by $V_{g}$. $V_{l}$ and $V_{g}$ are concatenated and sent through an expression classifier based on cross-entropy loss denoted by $L_{u}$. It is to be noted that the entire local and global context processing through SCAN is devoid of any information from external landmark detectors.

For the CCI branch,  the output of $Conv\_4x$ block from ResNet-50, denoted by $F_{l}$, goes as input. It has dimension  $14 \times 14 \times 1024$. As expressed in Eq. \ref{eq:3}, the output of CCI is a set of k vectors $\{ OF_{1}, OF_{2}, ..., OF_{k}\}$. Subsequently, each of  $OF_{i}$ is densely connected to a 256 node layer and then goes through an expression classifier based on cross-entropy loss, denoted by $L_{i}$. The total loss from the CCI branch is 
\begin{equation} \label{eq:4}
    L_{l} = \sum_{i=1}^{k} L_{i}
\end{equation}
\newline
The overall loss from both the branches is given by
\begin{equation}\label{eq:5}
    L = \lambda*L_{u} + (1-\lambda)*L_{l}
\end{equation}
where $\lambda$ belongs to [0, 1]. In this work, we set $\lambda$ to 0.2 based on the ablation study reported in Section \ref{subsec_ablation_studies}.
\section{Datasets}
\subsection{In-the-wild Datasets}
\textbf{AffectNet} \cite{12} is the largest facial expression dataset with around 0.4 million images manually labeled for the presence of eight (neutral, happy, angry, sad, fear, surprise, disgust, contempt) facial expressions along with the intensity of valence and arousal. Since some papers in literature quote results only for first seven expressions (without contempt) while few others quote for all eight expressions, to do a fair comparison, this work reports results on both the cases. The former case is denoted by \textbf{AffectNet-7} while the latter is denoted by \textbf{AffectNet-8}. Only maximum of 20000 images per expression are considered for training in both the cases. Test set sizes are 3500 and 4000 samples for both the cases,  respectively. Since the dataset has class imbalance, oversampling is used for training. Challenging test subsets of \textbf{AffectNet} namely \textbf{Occlusion-AffectNet} and \textbf{Pose-Affectnet} provided by \cite{22} are also considered. For the \textbf{Occlusion-Affectnet}, each image is occluded with at least one type of occlusion: wearing mask, wearing glasses etc. There are a total of 682 images. For the \textbf{Pose-AffectNet}, images with pose degrees larger than 30 and 45 are collected. The number of images are 1949 and 985, respectively. \textbf{RAF-DB} \cite{li2019reliable_45,li2017reliable_46} contains 29672 facial images tagged with basic or compound expressions by 40 independent taggers. In this work, only images with basic emotions are used, including 12271 images as training data and 3068 images as test data. Like in AffectNet, challenging test subsets of RAF-DB namely Occlusion-RAF-DB and Pose-RAF-DB given by \cite{22} are also considered. The \textbf{Occlusion-RAF-DB} has a total number of 735 images, while \textbf{Pose-RAF-DB} has 1248 and 558 images with pose larger than 30 and 45 degrees, respectively. \textbf{FERPlus} by \cite{47} consists of 28558 training images, 3589 validation images and 3589 test images. Each image is labeled with one of the eight expressions by 10 independent taggers.  Similar to AffectNet, challenging test subsets of \textbf{FERPlus} namely \textbf{Occlusion-FERPlus} and \textbf{Pose-FERPlus} \cite{22} are also considered. The \textbf{Occlusion-FERPlus} has a total number of 605 images, while \textbf{Pose-FERPlus} has 1171 and 634 images with pose larger than 30 and 45 degrees, respectively. \textbf{SFEW} \cite{49} contains selected frames from \textbf{AFEW} \cite{48}. It contains 891 training images and 431 validation images covering unconstrained facial expressions, different head poses, ages, and occlusions. \textbf{FED-RO} \cite{21} is a recently released facial expression dataset with real world occlusions. Each image has natural occlusions including sunglasses, medical mask, hands and hair. It contains 400 images labeled with seven expressions for testing. This dataset is used only for testing the model.
\subsection{In-lab Datasets}
The Extended Cohn-Kanade dataset \textbf{(CK+)} \cite{10_b, 10_a} contains 593 video sequences recorded from 123 subjects. The first and last three frames of each video sequence are considered as the neutral and target expressions, resulting in 1254 and 1308 images for the cases of 7 and 8 expressions, respectively. \textbf{Oulu-CASIA} \cite{11} dataset contains six prototypic expressions from 80 people between 23 to 58 years old. The first and the last three frames are considered as the neutral and target expressions, resulting in 1920 images. \textbf{JAFFE} \cite{44} contains 213 images of 7 facial expressions  posed by 10 Japanese female models. Each image has been rated on 6 emotion adjectives by 60 Japanese subjects.

\subsection{Synthetic Covid-19 mask Datasets}
    In the background of covid-19 pandemic where worldwide restrictions have enforced people to mask their nose and mouth regions, this work constructs two datasets from FERPlus and RAF-DB using the publicly available code\footnote{https://github.com/X-zhangyang/Real-World-Masked-Face-Dataset/tree/master/wear\_mask\_to\_face}, simulating masked faces in covid-19 scenario. Performance on these datasets are also reported.

\section{Implementation Details}
The proposed work is implemented in PyTorch\footnote{https://pytorch.org} using a single GeForce RTX 2080 Ti GPU with 11GB memory. The backbone net of the proposed architecture is a pre-trained ResNet-50, trained on VGGFace2 \cite{20}. Resnet-50 trained on ImageNet \cite{imagenet_cvpr09_50} is also considered. All images are aligned using MTCNN\footnote{https://github.com/ZhaoJ9014/face.evoLVe.PyTorch} \cite{zhang2016_56} and then resized to 224x224. Input to SCAN is output of Conv\_3x block of backbone Resnet-50.  Input to CCI is output of Conv\_4x block of backbone Resnet-50.  Model is optimized by stochastic gradient descent (SGD) for 60 epochs. Learning rate (lr) of backbone is initialized to 0.0001. With regard to SCAN and CCI, lr is initialized to 0.01 when trained on FERPlus and RAF-DB. For AffectNet, lr is initialized to 0.001. For SFEW training, the proposed model trained on combination of AffectNet and RAF-DB is fine tuned with lr set to 0.001 on SCAN and CCI branches. In case of in-lab datasets, lr for backbone, SCAN and CCI are all set to 0.00001 while performing 10-fold cross-validation. In all cases, lr for the backbone, SCAN and CCI branches are exponentially decayed by a factor of 0.95 every epoch. Momentum in SGD is set to 0.9 and weight-decay is set to 0.001. Batch size is fixed at 64. Evaluation metric used is overall accuracy. 
Since FERPlus has multiple labels per image, majority voting is adopted to choose the label. 

\section{Results and Discussions}\label{Resultsanddiscussion}
\subsection{Performance comparison on in-the-wild datasets with baselines}
The baselines considered are Resnet-50 trained from scratch (denoted here by Baseline1),  Resnet-50 pre-trained on ImageNet (denoted here by Baseline2) and Resnet-50 pre-trained on VGGFace2 (denoted here by Baseline3). Table \ref{tab:Table1} displays the comparison results.
\begin{table*}[hbt!]
    \centering
    \caption{Comparison of the proposed model against baselines on in-the-wild datasets}
    \begin{tabular}{c|c|c|c|c|c|c|c|c}
    \hline
    \multirow{2}{4em}{Model}  &  \multicolumn{2}{|c|}{FERPlus} & \multicolumn{2}{|c|}{RAFDB} & \multicolumn{2}{|c|}{AffectNet-7} & \multicolumn{2}{|c|}{SFEW} \\\cline{2-3} \cline{4-5}    \cline{6-7}\cline{8-9} 
    
    & Max & Mean$\pm$Std & Max & Mean$\pm$Std & Max & Mean$\pm$Std & Max & Mean$\pm$Std \\     
    \hline
    \hline
    Baseline1 & 55.82 & 53.04$\pm$2.04 & 59.87 & 50.99$\pm$7.25 & 45.43 & 44.24$\pm$1.33 & 21.34 & 21.02 $\pm$ 0.2 \\
    Baseline2 & 79.75 & 78.45$\pm$0.75 & 74.45 & 73.87$\pm$0.6 & 57.86 & 57.73$\pm$0.15 & 31.09 & 27.98 $\pm$ 2.64 \\
    Baseline3 & 79.31 & 78.79$\pm$0.46 & 76.43 & 76.25$\pm$0.16 & 56.77 & 56.3$\pm$0.44 & 20.18 & 19.63 $\pm$ 0.31 \\
    \textbf{Ours} & \textbf{89.42} & \textbf{89.13$\pm$0.25} & \textbf{89.02} & \textbf{88.50$\pm$0.16} & \textbf{65.14} & \textbf{64.7$\pm$0.28} & \textbf{58.93} &  \textbf{56.80$\pm$1.33} \\
    \hline
    \end{tabular}
    
    \label{tab:Table1}
\end{table*}

Clearly, the proposed model outperforms all the baselines with a minimum of around 9\% (in AffectNet-7) to a maximum of around 39\% (in SFEW) improvement over the best baseline. This indicates that pre-trained models from FR or ImageNet cannot be directly used for FER. Different branches (like SCAN and CCI) for processing feature maps from middle layers of  the pre-trained model are necessary for enhancing performance on FER.

\subsection{Comparison with state-of-the-art methods}
The proposed model is compared with GACNN\cite{21}, RAN\cite{22}, OADN\cite{23}, FMPN\cite{chen2019facial_51} and SCN\cite{wang2020_52}. These are the most recent state-of-the-art methods published in last one year. Table \ref{tab:Table2} reports the comparison of the proposed model against these methods on in-the-wild datasets. As can be seen, except the proposed method, none of the other 5 methods have reported results on all the datasets. From Table \ref{tab:Table2}, it is vivid that the proposed model has outperformed all the other models in all the datasets except RAF-DB. This clearly emphasizes that SCAN and CCI are vital for enhancing performance. It is to be noted that all the state-of-the-art methods have quoted only their best performance. For fair comparison, Table \ref{tab:Table2} highlights the best performance of our model across all the datasets. However, influence of randomness in training is clearly quantified for our model in the last row of Table \ref{tab:Table1}. Even while accounting for randomness, the one standard deviation interval of accuracy of the proposed model is sharper, emphasizing its superior generalizing capability. 

\begin{table}[hbt!]
    \centering
     \caption{Performance comparison against state-of-the-art-methods on in-the-wild datasets}
    \begin{tabular}{p{0.09\textwidth}|p{0.06\textwidth}|p{0.05\textwidth}|p{0.05\textwidth}|p{0.05\textwidth}|p{0.05\textwidth}}
         \hline
         Model & FERPlus & RAF-DB & Affect Net-7 & Affect Net-8 & SFEW  \\
         \hline
         \hline
         GACNN\cite{21}  & -     & 85.07     & 58.78     & -     & -  \\
         RAN\cite{22}    & 89.16 & 86.9      & -         & 59.5  & 56.4  \\
         OADN\cite{23}   & -     & \textbf{89.83}     & 64.06     & -     & -  \\
         FMPN\cite{chen2019facial_51}   & -     & -         & 61.52$^*$     & -     & -  \\
         SCN\cite{wang2020_52}   & 88.01  & 87.03     & -          & 60.23     & -  \\
         \textbf{Ours}   & \textbf{89.42}  & 89.02     & \textbf{65.14}          & \textbf{61.73}     & \textbf{58.93}  \\
         \hline 
        
         \multicolumn{6}{l}{$^*$\footnotesize{In FMPN, neutral images are not considered.}}
    \end{tabular}
   
    \label{tab:Table2}
\end{table}
\subsection{Performance on challenging test subsets of AffectNet, RAF-DB and FERPlus}
To validate the robustness of the proposed method to presence of occlusions and pose variations, the performance of the proposed method is tested on challenging test subsets of AffectNet, RAF-DB and FERPlus. It is also compared against RAN, and OADN. Table \ref{tab:Table3} depicts the results.
\begin{table*}[hbt!]
    \caption{Performance comparison on occlusions and pose variations subsets.}
    \centering
    \begin{tabular}{p{0.07\textwidth}|p{0.04\textwidth}|p{0.06\textwidth}|p{0.06\textwidth}|p{0.04\textwidth}|p{0.06\textwidth}|
                                            p{0.06\textwidth}|p{0.04\textwidth}|p{0.06\textwidth}|p{0.06\textwidth}|p{0.04\textwidth}|p{0.06\textwidth}|p{0.06\textwidth}}
         \hline
         \multirow{2}{4em}{Model} &  \multicolumn{3}{|c|}{FERPlus} & \multicolumn{3}{|c|}{RAFDB} & \multicolumn{3}{|c|}{AffectNet-7} & \multicolumn{3}{|c}{AffectNet-8}  \\ \cline{2-4} \cline{5-7} \cline{8-10}\cline{11-13}
         
          & Occ. & Pose$>$30 & Pose$>$45 &  Occ. & Pose$>$30 & Pose$>$45 &  Occ. & Pose$>$30 & Pose$>$45 &  Occ. & Pose$>$30 & Pose$>$45  \\
         \hline
         \hline
         RAN[22] & 83.63 & 82.23 & 83.63 &  82.72 & 86.74 & 85.2 & - & - & - & 58.5 & 53.9 & 53.19 \\
         OADN[23] & 84.57 & 88.52 & 87.50 &  - & - & - & 64.02 & 61.12 & 61.08 & - & - & - \\
         \textbf{Ours} & \textbf{86.12} & \textbf{88.89} & \textbf{88.15} &  \textbf{85.03} & \textbf{89.82} & \textbf{89.07} & \textbf{67.06} & \textbf{62.64} & 
                                                                             \textbf{61.31} & \textbf{63.69} & \textbf{60.41} & \textbf{60.86} \\
         \hline
    \end{tabular}
    \label{tab:Table3}
\end{table*}
The proposed method exhibits superior robustness to occlusions and pose variations, with around 2 to 5\%  gains in case of occlusions and around 1 to $7\%$ gains in case of pose variations, over RAN and OADN. To further validate the robustness to occlusions, the proposed method is tested on another real-world occlusions dataset, namely FED-RO. Table 4 enumerates the results. Again, the proposed method stands tall among all the state-of-the-art methods with a solid $2.3\%$ advantage over the next best performing OADN.
\begin{table}[hbt!]
    \begin{center}
    \caption{Performance comparison on FED-RO}
    \medskip
    \begin{tabular}{c|c}
         \hline
         Method & Average Accuracy \\
         \hline
         \hline
         GACNN\cite{21} & 66.5 \\
         RAN\cite{22} & 67.98 \\
         OADN\cite{23} & 71.17 \\
         \textbf{Ours} & \textbf{73.5} \\
         \hline
    \end{tabular}
    \end{center}
    \label{tab:Table4}
\end{table}

\subsection{Expression discrimination}
\begin{figure*}[hbt!]
    \centering
    \includegraphics[width=1.0\textwidth]{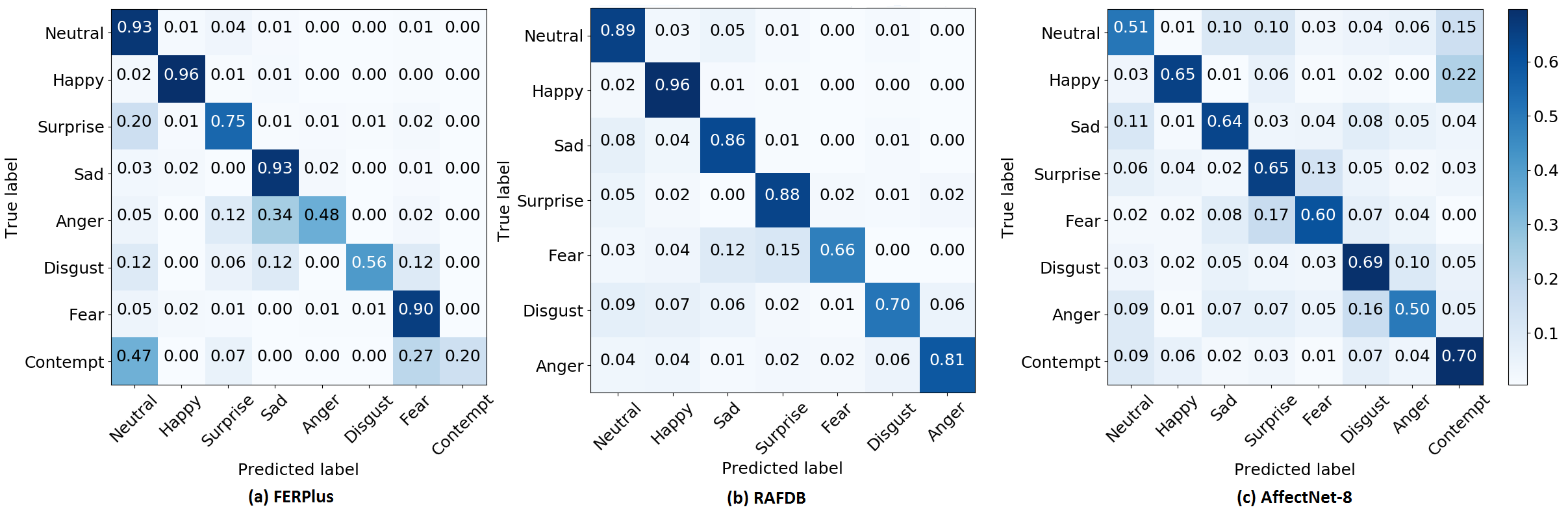}
    \caption{Confusion matrices for Ferplus, RAFDB and Affectnet\-8 datasets }
    \label{fig:figure_confusion_matrices}
\end{figure*}

In order to obtain a clear picture of easily identifiable expressions and most difficult expressions to distinguish, confusion matrices with regard to AffetNet-8, RAF-DB and FERPlus are plotted in Fig \ref{fig:figure_confusion_matrices}. More confusion matrices with regard to occlusions and pose variations across all the aforementioned datasets plus confusion matrices with regard to SFEW and FED-RO are presented in supplementary material. Happiness is the easiest expression to recognize in general except in AffectNet-8 where contempt and disgust are relatively easier expressions to recognize. Fear is generally confused with surprise and vice-versa except in FERPlus where fear is easily recognized along with happy expression. Presence of contempt expression in AffectNet-8 drastically brings down recall on happy expression. Another observation is that disgust is confused with anger sometimes, fear, sad and neutral other times. In general, negative expressions are confused with negative expressions though sometimes they are confused with happy (like disgust getting confused with happy in RAF-DB) or neutral (like contempt getting confused with neutral in FERPlus). Surprise can be happily surprised positive emotion or a negative emotion. The possible reasons for such confusion could be: (i) some of the negative emotions like disgust, contempt, sad are subtle to distinguish (ii) inconsistent annotation of expressions by humans. A workaround to these issues and thereby to further enhance the performance would be to rely on a hybrid approach that has an architecture like the proposed one in this paper and a meta-learner \cite{wang2020_52} that can suppress uncertainties in annotations automatically. This is a future work to be carried out.

\subsection{Performance on in-lab datasets}
Table \ref{tab:Table5} highlights the performance of the proposed method on CK+ and Oulu-CASIA datasets. 10-fold cross-validation is carried out. The proposed model trained on combination of AffectNet and RAF-DB is fine-tuned during the cross-validation. It is to be noted that there is no standard way of selection of expression images from the video sequences of CK+ and Oulu-CASIA. This work follows \cite{Ilke_54} in this regard. In Table \ref{tab:Table5}, CK+8 corresponds to consideration of all 8 expressions while CK+7 corresponds to only first 7 expressions (no contempt). Again, CK+7 for IACNN \cite{Meng_53}, DeRL \cite{40} and FMPN \cite{chen2019facial_51} corresponds to 7 expressions without contempt expression. 
\begin{table}[hbt!]
    \centering
    \caption{Performance comparison on in-lab datasets}
    \begin{tabular}{c|c|c|c}
         \hline
         Method    &   CK+8    &   CK+7    &   Oulu-CASIA \\
         \hline
         \hline
         Baseline1 &   91.72   &   -       &   82.71       \\
         IACNN \cite{Meng_53} &   -       &   95.37   &   -            \\
         DeRL \cite{40}  &   -       &   97.30   &   \textbf{88.0*}            \\
         FMPN \cite{chen2019facial_51}  &   -       &   \textbf{98.06}   &   -        \\
         Teacher-ExpNet \cite{Ilke_54}  & \textbf{97.6} &   -       &   85.83    \\
         \textbf{Ours}      &  \textbf{97.07}   & \textbf{97.31}  & \textbf{86.56} \\
         \hline
          \multicolumn{4}{l}{$^*$\footnotesize{Only six expressions without neutral is considered.}}
    \end{tabular}
    \label{tab:Table5}
\end{table}
The proposed method performs almost on par with the state-of-the-art methods for in-lab datasets. It is to be noted that most of the compared methods are specifically designed for in-lab datsets while the proposed method focuses on robustness to wild conditions like occlusions and pose variations in the real-world. Despite this difference, the proposed method performs well on in-lab datasets, which strengthens the generalizing capability of the proposed method. Apart from cross-validation through fine-tuning, cross-evaluation is also performed directly on in-lab datasets. For this, the best model trained on AffectNet-7 is evaluated on CK+, Oulu-CASIA and JAFFE datasets. Results are displayed in Table \ref{tab:Table6}. 
\begin{table}[hbt!]
    \centering
    \caption{Cross-evaluation performance of the proposed method}
    \begin{tabular}{c|c|c|c}
         \hline
         Model    &   CK+7    &   Oulu-CASIA    &   JAFFE \\
         \hline
         \hline
         GACNN\cite{21} &   91.64   &   58.18       &   -       \\
         Ours  &   91.04   &   \textbf{58.49}       &   56.33   \\
         \hline
    \end{tabular}
    
    \label{tab:Table6}
\end{table}

\subsection{Influence of SCAN} \label{subsection_SCAN}
To study how good SCAN is, the proposed model is trained on AffectNet by replacing SCAN using: (i) a single attention (SA), (ii)  a spatial attention constant across channels (SPA), (iii) only local patch attention (LPA), (iv) only global feature map attention (GA), (v) ECA. Table \ref{tab:Table7} elicits the comparison results. Definitely, SCAN has advantage over other attention modules. Though the gain may not look pronounced, it is to be noted that the local patches input to SCAN are of size around 5x5x512. The spatial size of 5x5 is small. However, a gain of around 0.3 to 1.2\% achieved by SCAN over other attention modules emphasize that even when spatial size is smaller, per channel per location attention pushes up the performance and makes the model more robust. It is also found in our experiment that absence of attention branch (absence of SCAN) brings down the performance of the model by 1.2 to 1.5\% across the datasets.
\begin{table}[hbt!]
    \centering
    \caption{Influence of SCAN}
    \begin{tabular}{c|c|c}
         \hline
         Attention Mechanism & AffectNet-7 & AffectNet-8 \\
         \hline
         \hline
         SA & 64.0 & 60.7 \\
         SPA & 64.43 & 61.35 \\
         LPA & 64.4 & 61.23 \\
         GA & 64.71 & 61.05 \\
         ECA & 64.54 & 61.18 \\
         \textbf{SCAN} & \textbf{65.14} & \textbf{61.73} \\
         \hline
    \end{tabular}
    \label{tab:Table7}
\end{table}

\subsection{Influence of CCI and ECA} \label{subsection_CCI_ECA}
To study how good is CCI branch, the proposed model is trained with and without CCI. The results are tabulated in Table \ref{tab:Table8}. It is very clear that without CCI, the performance drops down significantly, by 11 to 21\%. This emphasizes that wealth of information available from FR model must be put to use. In fact, in comparison to Table 1, it can be observed that a couple of baselines perform better than the proposed model without CCI. The reason is that, in the absence of CCI, the attention net SCAN alone cannot push the performance. SCAN operates on the output of the $Conv\_3x$ middle layer from the backbone Resnet-50 and it only provides attention. It does not provide representation. The representation comes from CCI which has a dense layer preceding the classifier layer. Whereas in the baselines, the representation learning/fine-tuning happens.

Further, weighting the features prior to use stresses important features and eliminates redundancies. To validate this point, Table \ref{tab:Table9} shows performance of the model in the presence and absence of ECA before CCI branch. Incorporating ECA before CCI branch does attain a significant gain of 2\% over ECA-less CCI branch.

\begin{table}[hbt!]
    \centering
    \caption{Influence of CCI}
    \begin{tabular}{c|c|c}
         \hline
          Database & Without CCI & With CCI \\
         \hline
         \hline
         AffectNet-7 & 43.74 & \textbf{65.14} \\
         AffectNet-8 & 42.63 & \textbf{61.73} \\
         RAF-DB & 73.79 & \textbf{89.02} \\
         FERPlus & 77.21 & \textbf{89.42} \\
         \hline
    \end{tabular}
    \label{tab:Table8}
\end{table}

\begin{table}[hbt!]
    \centering
    \caption{Influence of ECA}
    \begin{tabular}{c|c|c|c}
         \hline
            & AffectNet-7 & FERPlus & RAF-DB \\ 
         \hline
         \hline
         Without ECA & 63.51 &  88.71 & 86.34  \\
         With ECA & \textbf{65.14} &  \textbf{89.42} & \textbf{89.02} \\
         \hline
    \end{tabular}
    
    \label{tab:Table9}
\end{table}

\subsection{Performance on Masked Datasets Simulating Covid-19 Scenario}
\begin{figure}[ht!]
    \centering
    \includegraphics[width=0.6\textwidth]{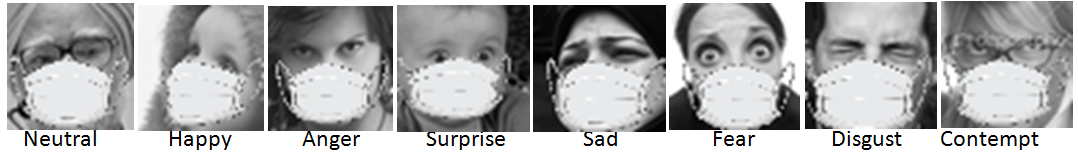}
    \caption{Example of synthetic masked face images for each of basic emotions from FERPlus dataset }
    \label{fig:figure_masked_faces}
\end{figure}

Another study carried out keeping in mind the current COVID-19 pandemic is to evaluate how well the proposed model perform when the regions of mouth and nose are masked. Of course, FED-RO is an already available real-world occlusions dataset on which the proposed method had already been evaluated (see Table 4). However, it has variety of occlusions and does not predominantly contain occlusions covering mouse and nose regions. The interest here is to look at the performance of the proposed model exclusively on masked mouth and nose regions. Towards this end, using the publicly available code \footnote{https://github.com/X-zhangyang/Real-World-Masked-Face-Dataset/tree/master/wear\_mask\_to\_face}, two datasets have been constructed from the whole of RAF-DB and FERPlus. A sample of constructed images are shown in Fig. \ref{fig:figure_masked_faces}.  Table \ref{tab:Table11} displays the results on these synthetic masked datasets.  Row 3 in Table \ref{tab:Table11} corresponds to the performance of the proposed model initialized by weights from the associated best performing proposed model on non-masked datasets. Similarly, row 4 corresponds to the performance of the proposed model initialized by weights from the associated best performing proposed model on non-masked datasets, and subsequently fine-tuned. Row 5 corresponds to the performance of the proposed model that is trained from scratch on masked datasets.  

Performance is compared with Baseline3 and RAN. Baseline1 and Baseline2 are not considered because they gave lower results than Baseline3. Compared to Baseline3, the proposed model has done exceedingly well in being robust to mask. Further, the proposed model has clearly outperformed RAN by around 3\% and 10\% on masked RAF-DB and masked FERPlus, respectively. However, a dip of around 13 to 14\% in performance is seen when compared to performance on non-masked datasets (see Table \ref{tab:Table1}). This is because some of the relevant regions like nose and mouth regions are unavailable for discriminating between expressions. Further, fine-tuning parameters of non-masked model does not perform as good as the one trained from scratch. Some kind of unlearning and fresh learning is required when certain regions are completely blocked. This is not possible in fine-tuning. This experiment clearly suggests that new ways need to be explored to tackle challenges as depicted here.

\begin{table}[hbt!]
    \centering
    \caption{Performance on masked RAF-DB and masked FERPlus datasets}
    \begin{tabular}{p{0.4\textwidth}|p{0.08\textwidth}|p{0.08\textwidth}}
         \hline
          Method                  & Masked RAF-DB   & Masked FERPlus  \\
         \hline
         \hline
         Baseline3               & 64.50           & 58.46     \\
         RAN(Trained from scratch) \cite{22}                    &  74.41                & 64.81          \\
         Ours(Pre-trained(Non-masked))  & 49.48           & 51.26     \\    
         Ours(Pre-trained(Non-masked)+ Finetuned)  & 75.65           & 71.6      \\
         Ours(Trained from scratch)     & \textbf{77.64}           & \textbf{74.69}     \\
         \hline 
    \end{tabular}
    \label{tab:Table11}
\end{table}

\subsection{Other Ablation Studies}\label{subsec_ablation_studies}
In SCAN, each local patch goes through a separate convolutional unit (see Eq.\ref{eq:1}). In this regard, an experiment is attempted where all the local patches share the convolutional parameters. The results are available in Table \ref{tab:Table10}. Clearly, shared parameters almost match the performance of SCAN with non-shared parameters. In fact, in RAF-DB, shared parameters outperform the non-shared SCAN. With shared parameters, downgrade by a factor of 25 in the number of parameters for SCAN can be achieved, making SCAN lighter.
\begin{table*}[hbt!]
    \centering
     \caption{Shared parameters vs non-shared parameters in SCAN}
    \begin{tabular}{c||c|c|c|c|c|c}
         \hline
         Model            & FERPlus & RAF-DB    & Affect Net-7 & Affect Net-8 & SFEW & FED-RO  \\
         \hline
         \hline
         Non-shared SCAN  & 89.42   & 89.02     &  65.14       & 61.73        & 58.93 & 73.5   \\
         Shared SCAN      & 89.25   & \textbf{89.11} & 64.43   & 60.47        & 56.61 & 72.75   \\
         \hline 
    \end{tabular}
   
    \label{tab:Table10}
\end{table*}

\begin{figure}[ht!]
    \centering
    \includegraphics[width=.6\textwidth, height =4cm]{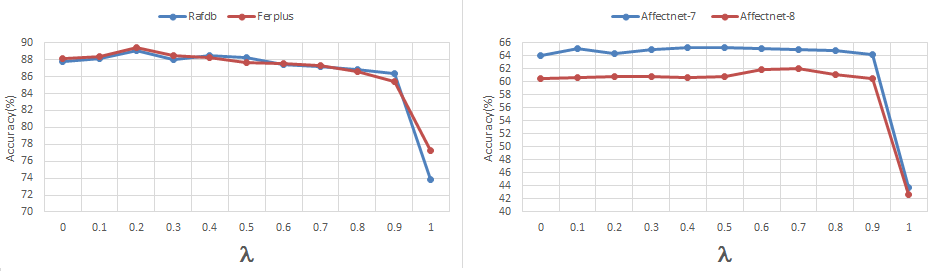}
    \caption{Performance for different values of the loss weighting parameter($\lambda$)  }
    \label{fig:figure_attwts}
\end{figure}

$\lambda$, the loss weighting parameter (see Eq.\ref{eq:5}) is another hyperparameter to be tuned. Plot in Fig. \ref{fig:figure_attwts}  depicts the performance against various values of $\lambda$. Setting  $\lambda=1$ is a worst option since the CCI branch becomes unavailable. Further, different  $\lambda$ values, between 0.2 and 0.5 gave best performances for different datasets. For example, in AffectNet-7, the model reached as high as \textbf{65.3} when  $\lambda$ equals 0.4. Similarly, in AffectNet-8,  the model reached as high as \textbf{61.93} when  $\lambda$ equals 0.6. RAF-DB and FERPlus attained their best performance when  $\lambda$ is 0.2. So, to present results consistently across all datasets, in this work $\lambda$ is set to 0.2.

\begin{figure}[ht!]
    \centering
    \includegraphics[width=.6\textwidth, height =4cm]{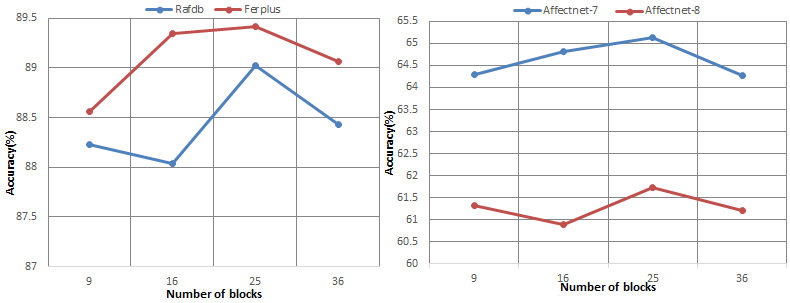}
    \caption{Performance for different number of blocks for partition }
    \label{fig:figure_blocks}
\end{figure}

Next, the number of local patches ‘m’ to consider in SCAN is a hyperparameter. Plot in Fig \ref{fig:figure_blocks} presents the performance against number of partition blocks in SCAN that define the number of local patches ‘m’. The margin of difference between different sizes is very thin. m=25 gave the best results.

\subsection{Visualizations}
\begin{figure}[hbt!]
    \centering
    \includegraphics[width=6cm, height=7cm]{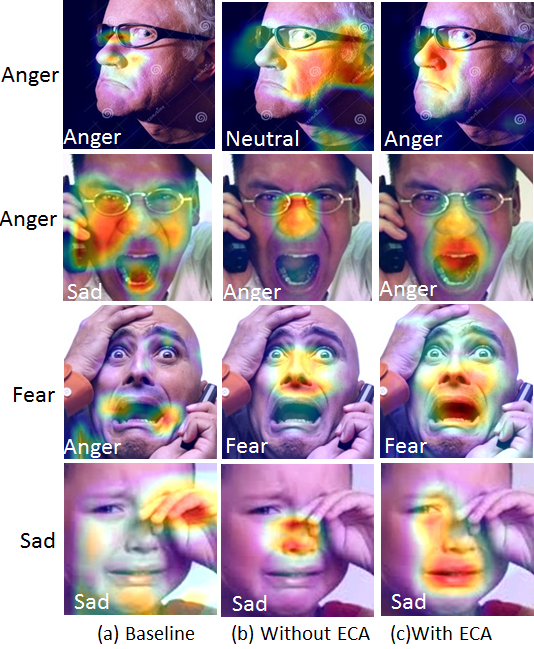}
    \caption{Comparison of activation maps captured for FED-RO images with and without ECA and Baseline3. True label is indicated on left (in black) and predicted label on bottom of each image(in white).}
    \label{fig:Visualizations using GRAD-CAM}
\end{figure}

     In order to investigate the salient regions focused upon by the proposed model in the presence of occlusions and pose variations, the attention weighted activation maps from $Conv\_3x$ are visualized using Grad-CAM\cite{Selvaraju_60} for Baseline3, the proposed model without ECA and the proposed model with ECA. 
      Fig. \ref{fig:Visualizations using GRAD-CAM} shows the visualizations. Dark color indicates high attention while lighter color indicates negligible attention.  For Baseline3, either the attention is spread over occluded parts (1st column, 2nd and 4th rows) or the attention is negligible all over. The proposed model without ECA also sometimes focuses on irrelevant pars (ear region in 2nd column, 1st row) or misses out on relevant parts (like missing out mouth region in 2nd to 4th rows). In comparison to Baseline3 and ECA-less case, the proposed model with ECA attends on non-occluded and relevant parts for expression recognition. Though the predictions made by Baseline3 and the proposed model without ECA are correct on some of the cases in Fig. \ref{fig:Visualizations using GRAD-CAM}, it is to be noted that presence of ECA provides attention that seems more natural. Also, the proposed model without ECA still has SCAN and the rest of the CCI branch to propel it to provide better attention weights than Baseline3.  More visualizations are provided in the supplementary material.
      
      Fig. \ref{fig:Failures_Visualizations} shows some of failure cases of the proposed method. It is to be noted that though the predictions are wrong, attention has been given to relevant regions (except may be in image 4), avoiding occluded parts. Wrong predictions are among confused pair of expressions. Usually, surprise is easily confused with happy and fear expressions and sad with disgust. Such confused pairs sometimes arise due to inconsistent labeling by human taggers, and also some expressions exhibiting compound emotions like happily surprised or fearfully sad. These failures open up more scope for research in FER domain in future.  
     
\begin{figure}[hbt!]
    \centering
    \includegraphics[width=0.5\textwidth]{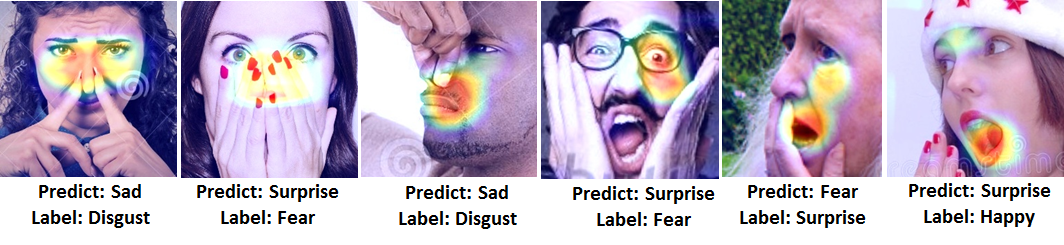}
    \caption{Some examples of failures of our method from FED-RO dataset }
    \label{fig:Failures_Visualizations}
\end{figure}

\section{Conclusions}
In conclusion, the efficacy of per channel per location attention coupled with complementary context information available through the knowledge transferred from middle layers of pre-trained FR model is demonstrated across both in-the-wild and in-lab datasets through a variety of experiments and illustrations in this work. Two directions to look into future are: (i) inconsistent annotations of FER datasets can bring down the performance. A hybrid approach of suppressing uncertainties in annotations combined an attention architecture similar to the one proposed in this work can further boost performance of FER. (ii) Recognizing expressions from faces masked in mouth and nose regions that is common in current COVID-19 pandemic requires definition of expressions based only on eyes and forehead regions. The feasibility of this and subsequent incorporation of these definitions into a learning paradigm is a useful direction to venture into.

\section*{Acknowledgments}
We dedicate this work to Bhagawan Sri Sathya Sai Baba, Divine Founder Chancellor of Sri Sathya Sai Institute of Higher Learning, PrasanthiNilyam, A.P., India.

\bibliographystyle{plain} 

\bibliography{main}

\begin{thebibliography}{10}

\bibitem{30}
Lanitis A., Taylor C., and Cootes T.F.
\newblock Automatic interpretation and coding of face images using flexible
  models.
\newblock {\em IEEE Transactions on Pattern Analysis and Machine Intelligence},
  pages 743--756, 2002.

\bibitem{1}
F.~{Abdat}, C.~{Maaoui}, and A.~{Pruski}.
\newblock Human-computer interaction using emotion recognition from facial
  expression.
\newblock In {\em 2011 UKSim 5th European Symposium on Computer Modeling and
  Simulation}, pages 196--201, 2011.

\bibitem{47}
E.~Barsoum, C.~Zhang, C.~C. Ferrer, and Z.~Zhang.
\newblock Training deep networks for facial expression recognition with
  crowdsourced label distribution.
\newblock {\em In Proceedings of the 18th ACM International Conference on
  Multimodal Interaction}, page 279–283, 2016.

\bibitem{41}
J.~Cai, Z.~Meng, A.~S. Khan, Z.~Li, J.~O’Reilly, and Y.~Tong.
\newblock Identity-free facial expression recognition using conditional
  generative adversarial network.
\newblock {\em arXiv preprint, arXiv:1903.08051}, 2019.

\bibitem{20}
Q.~Cao, L.~Shen, W.~Xie, O.~M. Parkhi, and A.~Zisserman.
\newblock Vggface2: A dataset for recognising face across pose and age.
\newblock {\em International Conference on Automatic Face and Gesture
  Recognition}, 2018.

\bibitem{7}
Shan CF, Gong SG, and McOwan PW.
\newblock Facial expression recognition based on local binary patterns: A
  comprehensive study.
\newblock {\em Image and Vision Computing}, 27:803--816, 2009.

\bibitem{chen2019facial_51}
Yuedong Chen, Jianfeng Wang, Shikai Chen, Zhongchao Shi, and Jianfei Cai.
\newblock Facial motion prior networks for facial expression recognition.
\newblock {\em IEEE Visual Communications and Image Processing (VCIP)}, 2019.

\bibitem{5}
Hok chun Lo and R.~Chung.
\newblock Facial expression recognition approach for performance animation.
\newblock {\em IEEE Proceedings Second International Workshop on Digital and
  Computational Video}, 6:613--622, Feb 2001.

\bibitem{Ilke_54}
Ilke Cugu, Eren Sener, and Emre Akbas.
\newblock Microexpnet: An extremely small and fast model for expression
  recognition from face images.
\newblock {\em arXiv preprint, arXiv:1711.07011}, 2019.

\bibitem{imagenet_cvpr09_50}
J.~Deng, W.~Dong, R.~Socher, L.-J. Li, K.~Li, and L.~Fei-Fei.
\newblock {ImageNet: A Large-Scale Hierarchical Image Database}.
\newblock In {\em CVPR09}, 2009.

\bibitem{48}
A.~Dhall, R.~Goecke, S.~Lucey, and T.~Gedeon.
\newblock Acted facial expressions in the wild database.
\newblock {\em In Technical Report}, 2011.

\bibitem{49}
A.~Dhall, R.~Goecke, S.~Lucey, and T.~Gedeon.
\newblock Static facial expression analysis in tough conditions: Data,
  evaluation protocol and benchmark.
\newblock {\em IEEE ICCV workshop BEFIT}, 2011.

\bibitem{23}
Hui Ding, Peng Zhou, and Rama Chellappa.
\newblock Occlusion-adaptive deep network for robust facial expression
  recognition.
\newblock {\em arXiv preprint, arXiv:2005.06040v1}, May 2020.

\bibitem{25}
X.~Dong, Y.~Yan, W.~Ouyang, and Y.~Yang.
\newblock Style aggregated network for facial landmark detection.
\newblock {\em CVPR}, page 379–388, 2018.

\bibitem{3}
Zixizng Fei, Erfu Yang, David Day-Uei, Stephen Butler, Winifred Ijomah, Xia Li,
  and Huiyu Zhou.
\newblock Deep convolution network based emotion analysis towards mental health
  care.
\newblock {\em Neurocomputing}, 388:212--227, 2020.

\bibitem{43}
Hiroshi Fukui, Tsubasa Hirakawa, Takayoshi Yamashita, and Hironobu Fujiyoshi.
\newblock Attention branch network: Learning of attention mechanism for visual
  explanation.
\newblock {\em CVPR}, 2019.

\bibitem{8}
Shih FY, Chuang CF, and Wang PSP.
\newblock Performance comparisons of facial expression recognition in jaffe
  database.
\newblock {\em International Journal of Pattern Recognition and Artificial
  Intelligence}, 22:445--459, 2008.

\bibitem{17}
Yandong Guo, Lei Zhang, Yuxiao Hu, Xiaodong He, and Jianfeng Gao.
\newblock Ms-celeb-1m: A dataset and benchmark for large-scale face
  recognition.
\newblock {\em ECCV}, 2016.

\bibitem{2}
Satori Hachisuka, Kenji Ishida, Takeshi Enya, and Masayoshi Kamijo.
\newblock Facial expression measurement for detecting driver drowsiness.
\newblock In Don Harris, editor, {\em Engineering Psychology and Cognitive
  Ergonomics}, pages 135--144, Berlin, Heidelberg, 2011. Springer Berlin
  Heidelberg.

\bibitem{39}
D.~Hamester, P.~Barros, and S.~Wermter.
\newblock Face expression recognition with a 2-channel convolutional neural
  network.
\newblock {\em IEEE International Joint Conference on Neural Networks}, pages
  1--8, 2015.

\bibitem{36}
P.~Hu, D.~Cai, S.~Wang, A.~Yao, and Y.~Chen.
\newblock Learning supervised scoring ensemble for emotion recognition in the
  wild.
\newblock {\em Proceedings of the 19th ACM International Conference on
  Multimodal Interaction.}, page 553–560, 2017.

\bibitem{27}
JihenKhalfallah and Jaleleddine Ben~Hadj Slama.
\newblock Facial expression recognition for intelligent tutoring systems in
  remote laboratories platform.
\newblock {\em Procedia Computer Science}, 73:274--281, 2015.

\bibitem{9}
Chen JK, Chen ZH, Chi ZR, and Fu~H.
\newblock Facial expression recognition based on facial components detection
  and hog features.
\newblock {\em Abstracts of scientific cooperations international workshops on
  electrical and computer engineering subfields}, page 64–69, 2014.

\bibitem{31}
Mase K.
\newblock Recognition of facial expressions for optical flow.
\newblock {\em IEICE Transactions on Special Issue on Computer Vision and Its
  Applications}, E74:3474–3483, 1991.

\bibitem{10_b}
T.~Kanade, J.~F. Cohn, and Y.~Tian.
\newblock Comprehensive database for facial expression analysis.
\newblock In {\em Proceedings of the Foruth IEEE International Conference on
  Automatic Face and Gesture Recognition(FG'OO)}, pages 46--53, Grenoble,
  France, 2000.

\bibitem{13}
A.~Krizhevsky, I.~Sutskever, and G.~Hinton.
\newblock Imagenet classification with deep convolutional neural networks.
\newblock {\em Advances in Neural Information Processing Systems}, 2012.

\bibitem{35}
G.~Levi and T.~Hassner.
\newblock Emotion recognition in the wild via convolutional neural networks and
  mapped binary patterns.
\newblock {\em Proceedings of the 2015 ACM on international conference on
  multimodal interaction}, page 503–510, 2015.

\bibitem{li2019reliable_45}
Shan Li and Weihong Deng.
\newblock Reliable crowdsourcing and deep locality-preserving learning for
  unconstrained facial expression recognition.
\newblock {\em IEEE Transactions on Image Processing}, 28(1):356--370, 2019.

\bibitem{li2017reliable_46}
Shan Li, Weihong Deng, and JunPing Du.
\newblock Reliable crowdsourcing and deep locality-preserving learning for
  expression recognition in the wild.
\newblock In {\em 2017 IEEE Conference on Computer Vision and Pattern
  Recognition (CVPR)}, pages 2584--2593. IEEE, 2017.

\bibitem{21}
Yong Li, Jiabei Zeng, Shiguang Shan, and Xilin Chen.
\newblock Occlusion aware facial expression recognition using cnn with
  attention mechanism.
\newblock {\em IEEE Transactions on Image Processing}, 28:2439--2450, May 2019.

\bibitem{10_a}
P.~Lucey, J.~F. Cohn, T.~Kanade, J.~Saragih, Z.~Ambadar, and I.~Matthews.
\newblock The extended cohn-kanade dataset (ck+): A complete dataset for action
  unit and emotion-specified expression.
\newblock In {\em Computer Vision and Pattern Recognition Workshops(CVPRW)},
  page 94–101, San Francisco, USA, 2010.

\bibitem{44}
Michael~J. Lyons, Shigeru Akamatsu, Miyuki Kamachi, and Jiro Gyoba.
\newblock Coding facial expressions with gabor wavelets.
\newblock {\em 3rd IEEE International Conference on Automatic Face and Gesture
  Recognition}, pages 200--205, 1998.

\bibitem{6}
Jaffar MA.
\newblock Facial expression recognition using hybrid texture features based
  ensemble classifier.
\newblock {\em International Journal of Advanced Computer Science and
  Applications(IJACSA)}, 6, 2017.

\bibitem{33}
M.~Matsugu, K.~Mori, Y.~Mitari, and Y.~Kaneda.
\newblock Subject independent facial expression recognition with robust face
  detection using a convolutional neural network.
\newblock {\em Neural Networks}, 16:555–559, 2003.

\bibitem{Meng_53}
Z.~Meng, P.~Liu, J.~Cai, S.~Han, and Y.~Tong.
\newblock Identity-aware convolutional neural network for facial expression
  recognition.
\newblock {\em FG}, pages 558--565, 2017.

\bibitem{12}
Ali Mollahosseini, Behzad Hasani, and Mohammad~H Mahoor.
\newblock Affectnet: A database for facial expression, valence, and arousal
  computing in the wild.
\newblock {\em IEEE Transactions on Affective Computing}, 2017.

\bibitem{4}
Udayakumar N.
\newblock Facial expression recognition system for autistic children in virtual
  reality environment.
\newblock {\em International Journal of Scientific and Research Publications},
  6:613--622, Jun2016.

\bibitem{34}
Emilio~Soria Olivas, Jose David~Martin Guerrero, Marcelino~Martinez Sober, Jose
  Rafael~Magdalena Benedito, and Antonio Jose~Serrano Lope.
\newblock {\em Handbook Of Research On Machine Learning Applications and
  Trends: Algorithms, Methods and Techniques}.
\newblock IGI Publishing, 701, E. Chocolate Avenue, Suite 200, Hershey PA,
  2009.

\bibitem{19}
O.~M. Parkhi, A.~Vedaldi, and A.~Zisserman.
\newblock Deep face recognition.
\newblock {\em British Machine Vision Conference}, 2015.

\bibitem{14}
Olga Russakovsky, Jia Deng, Jonathan~Krause Hao~Su, Sanjeev Satheesh, Sean Ma,
  Zhiheng Huang, Andrej Karpathy, Aditya Khosla, Michael Bernstein,
  Alexander~C. Berg, and Li~Fei-Fei.
\newblock Imagenet large scale visual recognition challenge.
\newblock {\em International Journal of Computer Vision}, 115:211--252, Dec
  2015.

\bibitem{Selvaraju_60}
R.~R. Selvaraju, M.~Cogswell, A.~Das, R.~Vedantam, D.~Parikh, , and D.~Batra.
\newblock Grad-cam: Visual explanations from deep networks via gradient-based
  localization.
\newblock {\em https://arxiv.org/abs/1610.02391}, 2016.

\bibitem{15}
Karen Simonyan and Andrew Zisserman.
\newblock Very deep convolutional networks for large-scale image recognition.
\newblock {\em International Conference on Learning Representations}, 2015.

\bibitem{29}
Lingxue Song, Dihong Gong, Zhifeng Li, Changsong Liu, and Wei Liu.
\newblock Occlusion robust face recognition based on mask learning with
  pairwise differential siamese network.
\newblock {\em ICCV}, 2019.

\bibitem{37}
N.~Sun, Q.~Li, R.~Huan, J.~Liu, and G.~Han.
\newblock Deep spatial-temporal feature fusion for facial expression
  recognition in static images.
\newblock {\em Pattern Recognition Letters}, 2017.

\bibitem{16}
Christian Szegedy, Wei Liu, Yangqing Jia, Pierre Sermanet, Scott Reed, Dragomir
  Anguelov, Dumitru Erhan, Vincent Vanhoucke, , and Andrew Rabinovich.
\newblock Going deeper with convolutions.
\newblock {\em IEEE Conference on Computer Vision and Pattern Recognition
  (CVPR)}, June 2015.

\bibitem{42}
Nian~Chi Tay, Connie Tee, Thian~Song Ong, and Pin~Shen Teh.
\newblock Abnormal behavior recognition using cnn-lstm with attention
  mechanism.
\newblock {\em IEEE 1st International Conference on Electrical, Control and
  Instrumentation Engineering}, 2019.

\bibitem{wang2020_52}
Kai Wang, Xiaojiang Peng, Jianfei Yang, Shijian Lu, and Yu~Qiao.
\newblock Suppressing uncertainties for large-scale facial expression
  recognition.
\newblock In {\em 2020 IEEE Conference on Computer Vision and Pattern
  Recognition (CVPR)}. IEEE, 2020.

\bibitem{22}
Kai Wang, Xiaojiang Peng, Jianfei Yang, Debin Meng, and Yu~Qiao.
\newblock Region attention networks for pose and occlusion robust facial
  expression recognition.
\newblock {\em IEEE Transactions on Image Processing}, 29:4057-- 4069, January
  2020.

\bibitem{26}
Qilong Wang, Banggu Wu, Pengfei Zhu, Peihua Li, Wangmeng Zuo, and Qinghua Hu.
\newblock Eca-net: Efficient channel attention for deep convolutional neural
  networks.
\newblock {\em arXiv preprint, arXiv:1910.03151v4}, April 2020.

\bibitem{32}
Shinohara Y. and Otsu N.
\newblock Facial expression recognition using fisher weight maps.
\newblock {\em Proceedings of the Sixth IEEE Conference on Automatic Face and
  Gesture Recognition}, page 499–504, 2004.

\bibitem{40}
H.~Yang, U.~Ciftci, and L.~Yin.
\newblock Facial expression recognition by de-expression residue learning.
\newblock {\em CVPR}, page 2168–2177, 2018.

\bibitem{38}
A.~Yao, D.~Cai, P.~Hu, S.~Wang, L.~Sha, and Y.~Chen.
\newblock Holonet: towards robust emotion recognition in the wild.
\newblock {\em Proceedings of the 18th ACM International Conference on
  Multimodal Interaction}, page 472–478, 2016.

\bibitem{18}
D.~Yi, Z.~Lei, S.~Liao, and S.~Z. Li.
\newblock Learning face representation from scratch.
\newblock {\em arXiv preprint, arXiv:1411.7923}, 2014.

\bibitem{28}
J.~Yu, Z.~Lin, J.~Yang, X.~Shen, X.~Lu, and T.~S. Huang.
\newblock Generative image inpainting with contextual attention.
\newblock {\em CVPR}, 2018.

\bibitem{24}
J.~Zhang, M.~Kan, S.~Shan, and X.~Chen.
\newblock Occlusion-free face alignment: Deep regression networks coupled with
  de-corrupt autoencoders.
\newblock {\em CVPR}, page 3428–3437, June 2016.

\bibitem{zhang2016_56}
K.~Zhang, Z.~Zhang, Z.~Li, and Y.~Qiao.
\newblock Joint face detection and alignment using multitask cascaded
  convolutional networks.
\newblock {\em IEEE Signal Processing Letters}, 23(10):1499--1503, 2016.

\bibitem{11}
G.~Zhao, X.~Huang, M.~Taini, S.~Z. Li, and M.~Pietikainen.
\newblock Facial expression recognition from near-infrared videos.
\newblock {\em Image and Vision Computing}, 29:607--619, 2011.

\end{thebibliography}

\end{document}


\section{Supplementary material}
\subsection{Effect of oversampling}
In-the-wild datasets are highly imbalanced as it is difficult to annotate and collect images in categories like disgust and contempt. So, oversampling is used to overcome it. Results on AffectNet-7 with and without oversampling are shown in the Table \ref{tab:Table12} below:
\begin{table}[H]
    \centering
     \caption{Performance on AffectNet-7 with and without oversampling}
    \begin{tabular}{c|c}
         \hline
          Dataset      &    Accuracy \\
         \hline
         \hline
         With-oversampling & 65.14 \\
         \hline
         Without-oversampling & 62.57 \\
         \hline 
    \end{tabular}
   
    \label{tab:Table12}
\end{table}

\subsection{Inference time}
 The average inference time per 224 x 224 RGB image on the proposed model is calculated on GPU as well as on CPU. These are displayed in the Table \ref{tab:Table13} below:
\begin{table}[H]
    \centering
     \caption{Average inference time per image}
    \begin{tabular}{c|c}
         \hline
                &    Average time/image(sec) \\
         \hline
         \hline
         GPU & 0.03 \\
         \hline
         CPU & 0.12 \\
         \hline 
    \end{tabular}
   
    \label{tab:Table13}
\end{table}

\subsection{Visualizations}
\begin{figure}[H]
    \centering
    \includegraphics[width=.3\textwidth,height=6cm]{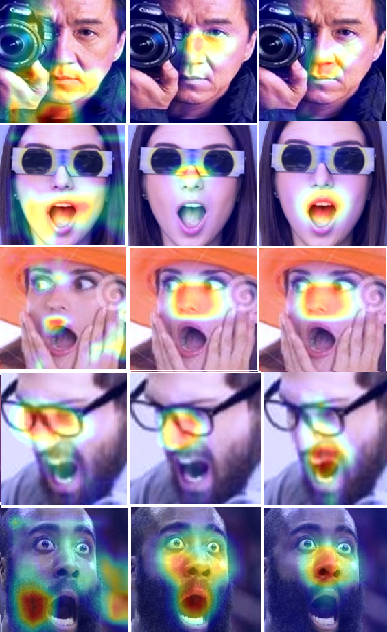}
    \caption{Activation Maps for FEDRO dataset images using Baseline3(left), our model without ECA(middle) and with ECA(right).}
    \label{fig:Activation_maps_extra}
\end{figure}

\begin{figure}[hbt!]
    \centering
    \includegraphics[width=.5\textwidth,height=6cm]{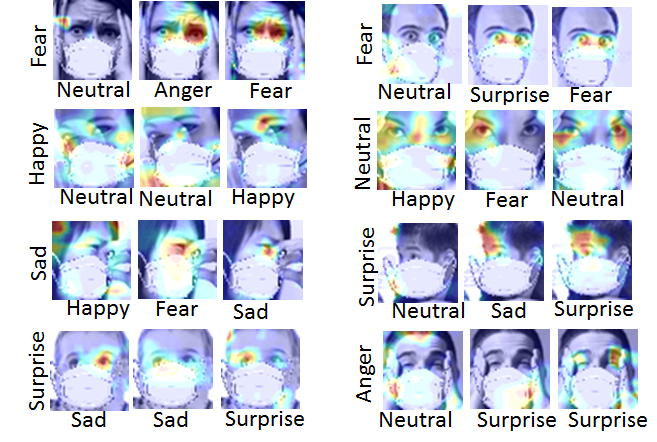}
    \caption{Comparison of activation maps on our masked FERPlus dataset between Baseline3 (left image in each triplet), RAN(middle image in each triplet) and our model (right image in each triplet). Image labels are displayed on extreme left of each pair. Predicted labels are below images.  }
    \label{fig:masked_activation_maps_extra}
\end{figure}

Additional visualizations, continuing from main text, are provided in Fig. \ref{fig:Activation_maps_extra}. Further, visualizations, as described in main text, are also provided for masked FERPlus dataset in Fig. \ref{fig:masked_activation_maps_extra}. It is clear that the proposed model did not focus on masked area while baseline and RAN spreads around occluded area also. 
\subsection{Confusion plots}
More confusion plots for different datasets are available in Figures [3-8]. Happy expression is easily recognizable across most of the datasets. Contempt expression could not be recognized well in FERPlus as number of samples for this category is much less as compared to others. Even though oversampling has been used, it requires further enquiry how performance can be improved. In SFEW dataset, disgust and anger are poorly recognized, again probably because of imbalance. Fear and surprise are confused pair of expressions. Similarly, anger and sad are confused at times.  

\begin{figure}[H]
    \centering
    \includegraphics[width=.6\textwidth,height=5cm]{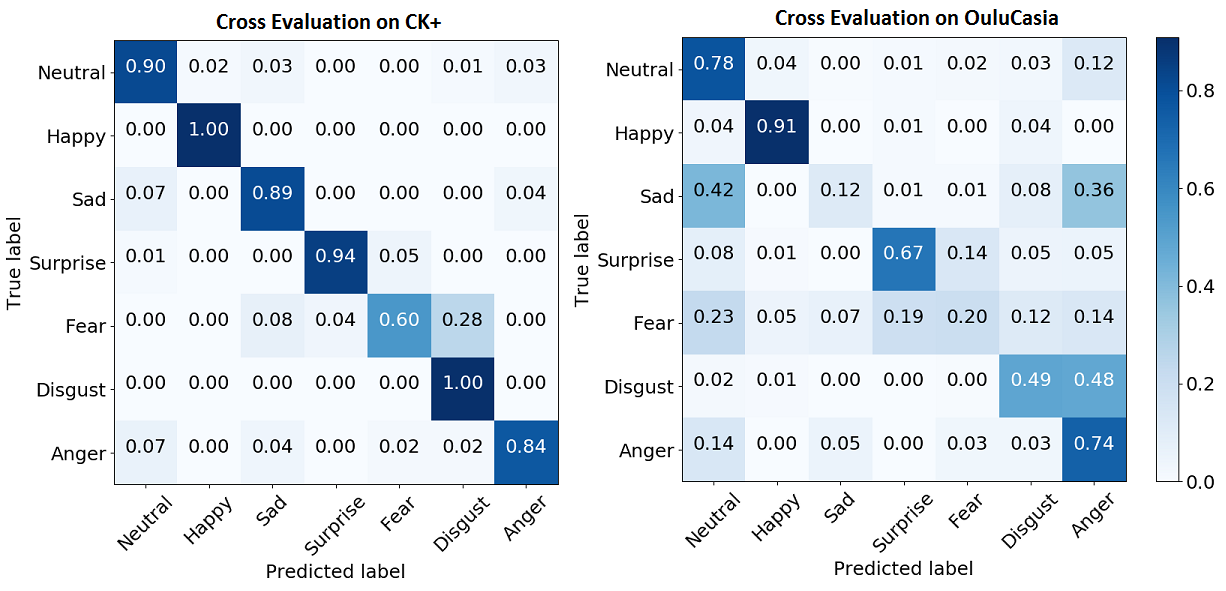}
    \caption{Confusion plots using cross-evaluation for CK+(7\-classes) and OuluCASIA datasets }
    \label{fig:Ckplus_oulucasia_confusion_plots}
\end{figure}

\begin{figure}[H]
    \centering
    \includegraphics[width=.6\textwidth,height=5cm]{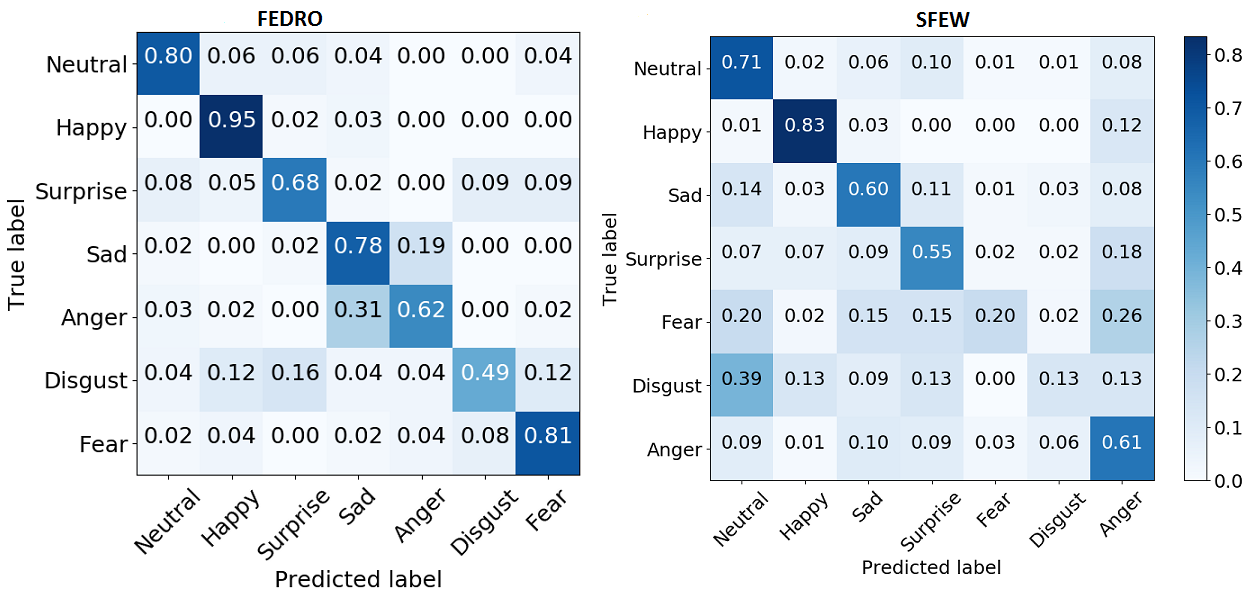}
    \caption{Confusion plots for SFEW and FED-RO datasets}
    \label{fig:sfew_fedro_confusion_plots}
\end{figure}

\begin{figure*}[ht!]
    \centering
    \includegraphics[width=1.0\textwidth,height=6cm]{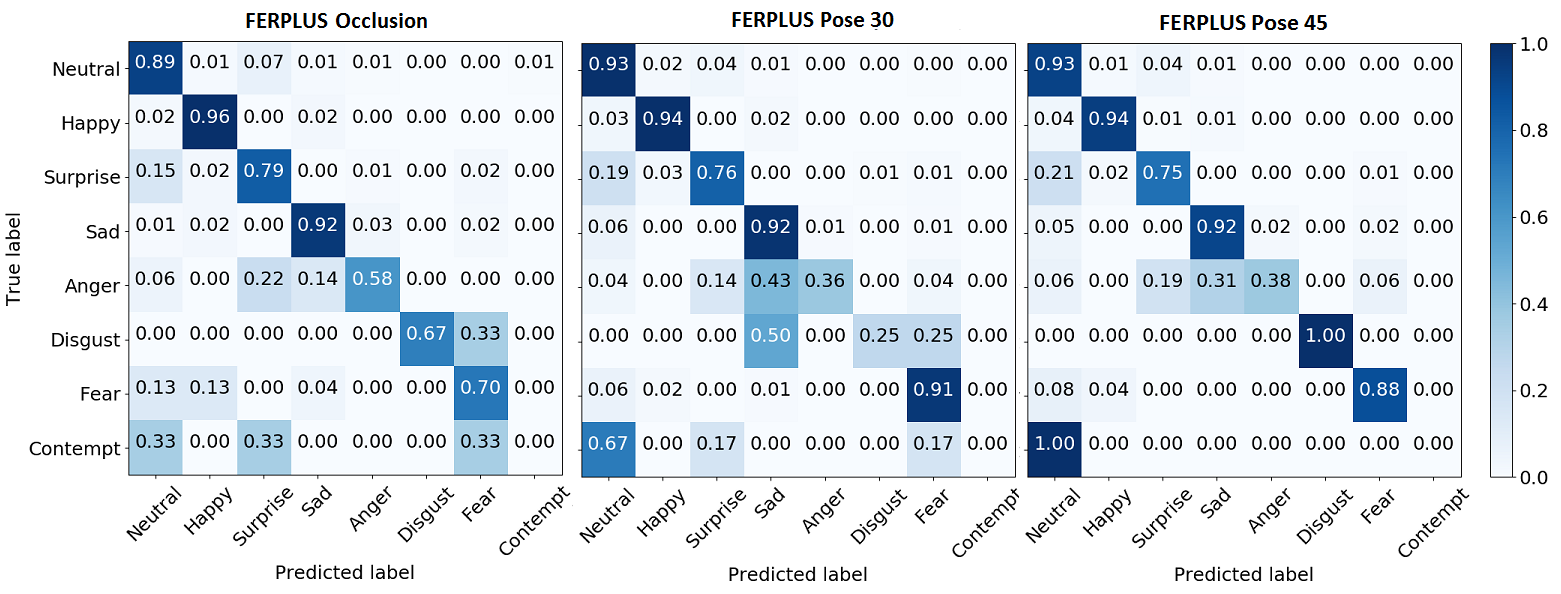}
    \caption{Confusion plots for Occlusion, Pose30 and Pose45 subsets of FERPLUS}
    \label{fig:FERPLUS_Confusionplots}
\end{figure*}

\begin{figure*}[ht!]
    \centering
    \includegraphics[width=1.0\textwidth,height=6cm]{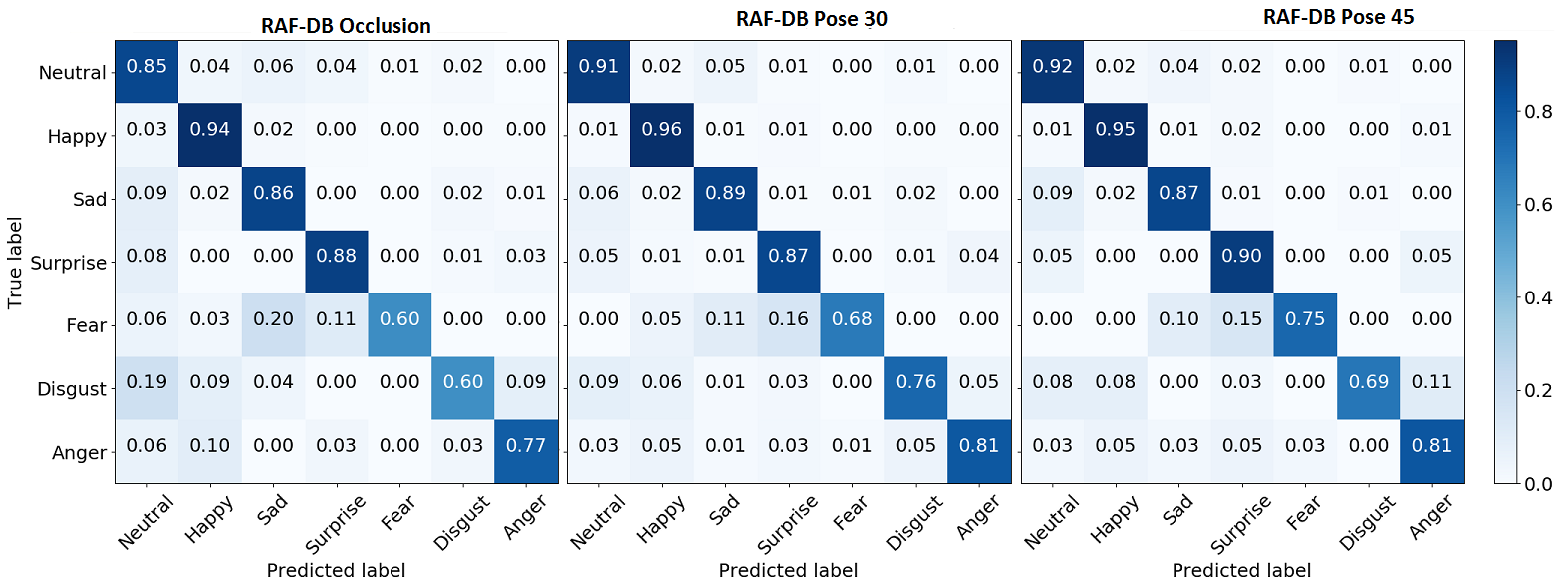}
    \caption{Confusion plots for Occlusion, Pose30 and Pose45 subsets of RAFDB}
    \label{fig:RAFDB_Confusionplots}
\end{figure*}

\begin{figure*}[ht!]
    \centering
    \includegraphics[width=1.0\textwidth,height=6cm]{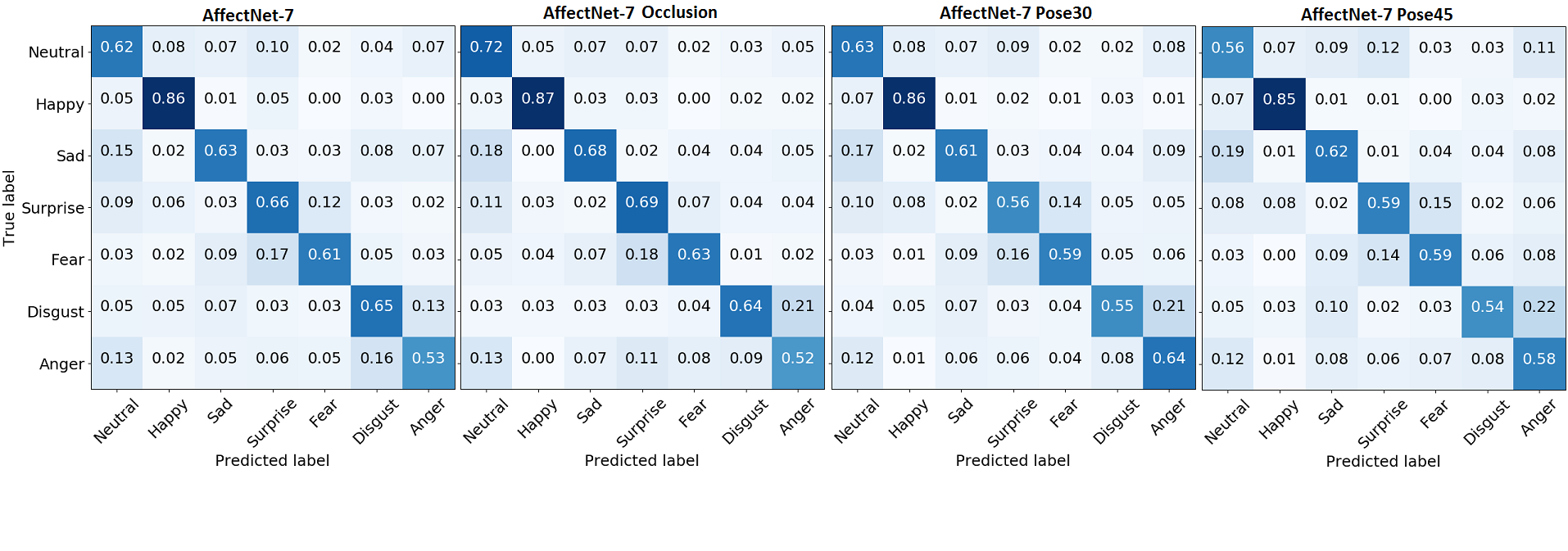}
    \caption{Confusion plots for Occlusion, Pose30 and Pose45 subsets of Affectnet on 7 classes}
    \label{fig:Affectnet7_Confusionplots}
\end{figure*}

\begin{figure*}[ht!]
    \centering
    \includegraphics[width=1.0\textwidth,height=6cm]{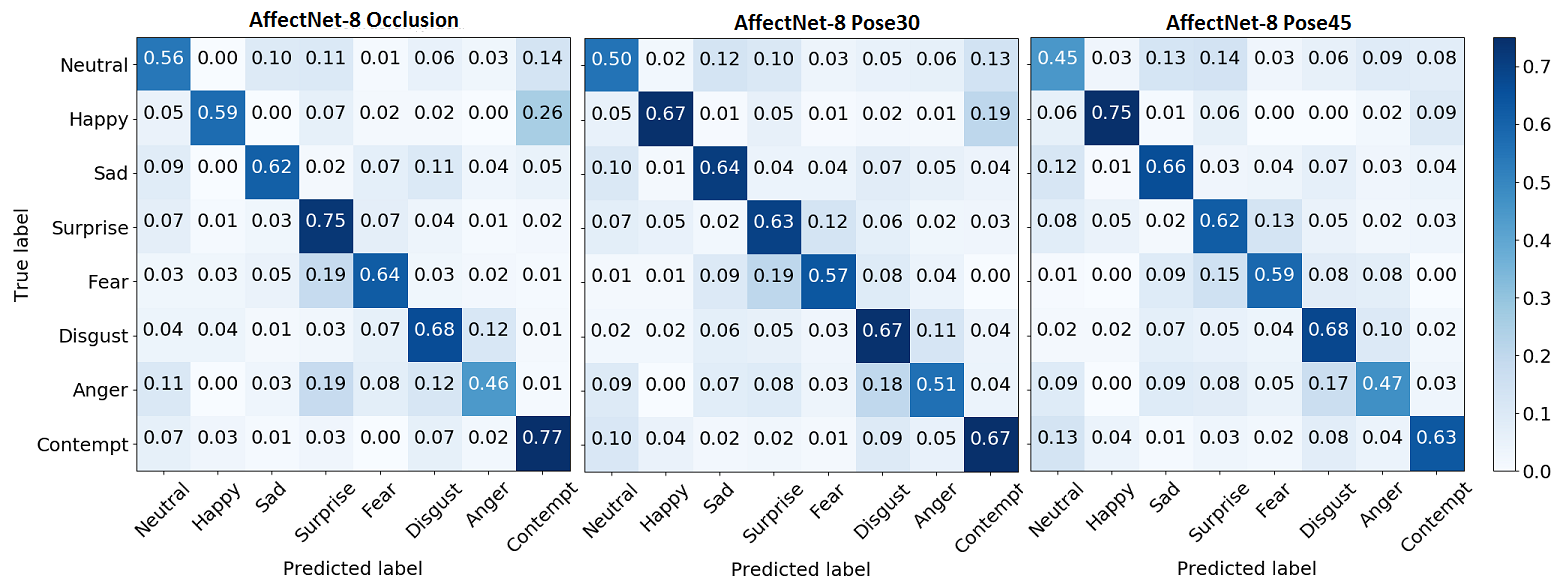}
    \caption{Confusion plots for Occlusion, Pose30 and Pose45 subsets of Affectnet on 8 classes}
    \label{fig:Affectnet8_Confusionplots}
\end{figure*}